\documentclass[times, review, 10pt]{elsarticle}

\usepackage{setspace}
\usepackage{amsmath,amssymb,amsfonts}
\usepackage{algorithmic}
\usepackage{graphicx}
\usepackage{textcomp}
\usepackage{booktabs} 
\usepackage{multirow}
\usepackage{caption}
\usepackage{url}

\begin{document}

\begin{frontmatter}

\title{\fontsize{14pt}{16pt}\selectfont ReHARK: Refined Hybrid Adaptive RBF Kernels for Robust One-Shot Vision-Language Adaptation}

\author{Md Jahidul Islam}
\ead{2006123@eee.buet.ac.bd}

\affiliation{organization={Department of Electrical and Electronic Engineering},
            addressline={Bangladesh University of Engineering and Technology}, 
            city={Dhaka},
            country={Bangladesh}}

\begin{abstract}
\fontsize{10pt}{12pt}\selectfont
\doublespacing
The adaptation of large-scale Vision-Language Models (VLMs) like CLIP to downstream tasks with extremely limited data—specifically in the one-shot regime—is often hindered by a significant ``Stability-Plasticity'' dilemma. While efficient caching mechanisms have been introduced by training-free methods such as Tip-Adapter, these approaches often function as local Nadaraya-Watson estimators. Such estimators are characterized by inherent boundary bias and a lack of global structural regularization. In this paper, \textbf{ReHARK} (\textbf{Re}fined \textbf{H}ybrid \textbf{A}daptive \textbf{R}BF \textbf{K}ernels) is proposed as a synergistic training-free framework that reinterprets few-shot adaptation through global proximal regularization in a Reproducing Kernel Hilbert Space (RKHS). A multi-stage refinement pipeline is introduced, consisting of: (1) \textbf{Hybrid Prior Construction}, where zero-shot textual knowledge from CLIP and GPT3 is fused with visual class prototypes to form a robust semantic-visual anchor; (2) \textbf{Support Set Augmentation (Bridging)}, where intermediate samples are generated to smooth the transition between visual and textual modalities; (3) \textbf{Adaptive Distribution Rectification}, where test feature statistics are aligned with the augmented support set to mitigate domain shifts; and (4) \textbf{Multi-Scale RBF Kernels}, where an ensemble of kernels is employed to capture complex feature geometries across diverse scales. Superior stability and accuracy are demonstrated through extensive experiments on 11 diverse benchmarks. A new state-of-the-art for one-shot adaptation is established by ReHARK, which achieves an average accuracy of \textbf{65.83\%}, significantly outperforming existing baselines. Code is available at \url{https://github.com/Jahid12012021/ReHARK}.
\end{abstract}

\begin{keyword}
Vision-Language Models \sep One-Shot Learning \sep Kernel Ridge Regression \sep GPT3 Semantics \sep CLIP.
\end{keyword}

\end{frontmatter}

\section{Introduction}
Vision-Language Models (VLMs), exemplified by CLIP~\cite{radford2021learning} and ALIGN~\cite{jia2021scaling}, have fundamentally reshaped the landscape of computer vision. By pre-training on billion-scale datasets of noisy image-text pairs via contrastive learning, these models align visual and semantic representations in a unified embedding space. This alignment grants them unprecedented zero-shot generalization capabilities, allowing them to recognize arbitrary concepts without task-specific training~\cite{yuan2021florence}. However, despite their robustness, deploying VLMs in downstream applications often requires adaptation to specific domains where the pre-training distribution differs significantly from the target distribution~\cite{zhang2022tip, bendou2025proker}.

Adapting these large-scale models with limited data—a setting known as \textit{few-shot learning}—presents a formidable ``Stability-Plasticity'' dilemma~\cite{zhou2022learning}. While fine-tuning-based approaches like CoOp~\cite{zhou2022learning} and CLIP-Adapter~\cite{gao2023clip} offer high performance, they are often computationally prohibitive and prone to \textit{catastrophic forgetting}~\cite{wortsman2022robust}. Conversely, training-free methods such as Tip-Adapter~\cite{zhang2022tip} have gained attention for their lightweight adaptation without the need for additional fine-tuning. Tip-Adapter utilizes a query-key cache model constructed from the few-shot training set, effectively retrieving few-shot knowledge in a non-parametric manner.

Despite its efficiency, recent theoretical analysis has revealed that Tip-Adapter functions as a modified version of the Nadaraya-Watson (NW) estimator—a local nonparametric regression method~\cite{bendou2025proker}. This local approach is known to suffer from significant boundary bias, which limits its ability to capture global task structures~\cite{hastie2009elements}. To mitigate these limitations, ProKeR~\cite{bendou2025proker} introduced a global adaptation method that learns a proximal regularizer in a reproducing kernel Hilbert space (RKHS). While ProKeR provides a more effective way to preserve prior knowledge, its performance in the extremely data-scarce \textit{one-shot regime} remains constrained by the difficulty of capturing domain-specific nuances from a single visual example.

\begin{figure}[t]
    \centering
    \includegraphics[width=\linewidth]{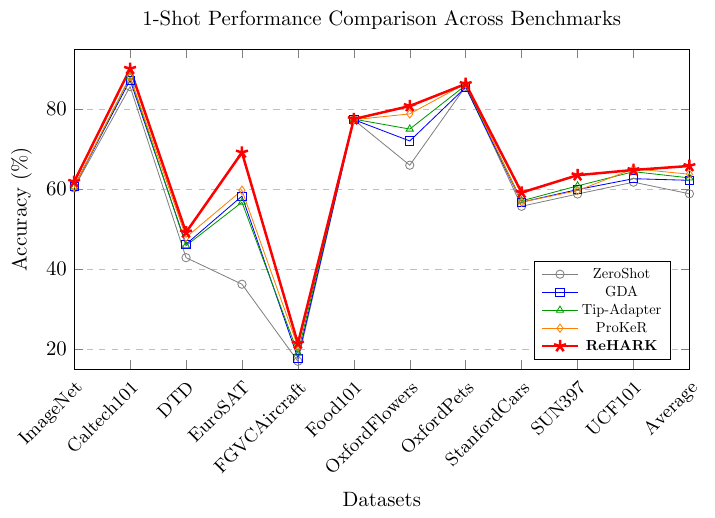}
    \caption{\doublespacing
    1-Shot performance comparison across 11 benchmarks. The proposed \textbf{ReHARK} method (red line with star markers) consistently outperforms existing training-free adaptation baselines.}
    \label{fig:performance_line_plot}
\end{figure}

In this work, \textbf{ReHARK} (Refined Hybrid Adaptive RBF Kernels) is introduced as a unified framework that resolves these issues by encoding multi-modal inductive biases and global regularization directly into the adaptation architecture. Several critical innovations are utilized:

\begin{enumerate}
    \item \textbf{Hybrid Semantic-Visual Prior Refinement:} It is argued that 1-shot visual evidence alone is insufficient for robust adaptation. A synergistic prior is constructed by blending CLIP text weights, high-density GPT3 semantic descriptions, and visual class prototypes. This hybrid prior stabilizes the global anchor of the model against domain-specific noise.
    \item \textbf{Adaptive Distribution Rectification and Bridging:} To resolve discrepancies between support and query distributions, a non-linear power transform and a distribution rectification step are applied to align test statistics with the training data. Furthermore, a ``Bridge'' mechanism is introduced to generate intermediate support samples by blending visual features with refined textual priors, effectively smoothing the adaptation manifold.
    \item \textbf{Ensemble Multi-Scale RBF Kernels:} Recognizing that a single kernel bandwidth is rarely optimal across diverse datasets, a Multi-Scale RBF kernel ensemble is utilized. By adaptively mixing kernels with different bandwidths, complex feature geometries across local and global scales are captured, which is critical for handling the high variance inherent in one-shot learning.
\end{enumerate}

\section{Related Work}

\subsection{Vision-Language Models and Zero-Shot Learning}
The landscape of computer vision has been fundamentally reshaped by the emergence of large-scale Vision-Language Models (VLMs) such as CLIP~\cite{radford2021learning} and ALIGN~\cite{jia2021scaling}. By performing contrastive pre-training on billion-scale image-text pairs, these models align visual and semantic representations within a unified embedding space~\cite{radford2021learning}. Such alignment facilitates unprecedented zero-shot generalization, enabling the recognition of arbitrary categories without the requirement for task-specific training data~\cite{yuan2021florence}. However, while zero-shot robustness is maintained across broad domains, performance is often found to be suboptimal when significant distribution shifts occur between pre-training and target datasets~\cite{zhang2022tip, bendou2025proker}.

\subsection{Few-Shot Adaptation and Prompt Learning}
To enhance the downstream performance of VLMs, various few-shot adaptation techniques have been developed. These are generally divided into parameter-efficient fine-tuning (PEFT) and training-free approaches. Among PEFT methods, prompt learning, exemplified by CoOp~\cite{zhou2022learning}, optimizes continuous learnable vectors in the text encoder's input space. Deep multimodal alignment is further pursued by methods like MaPLe~\cite{khattak2023maple}, which injects learnable tokens into both the vision and language branches. Alternatively, adapter-based methods such as CLIP-Adapter~\cite{gao2023clip} insert lightweight residual MLP modules into the frozen backbone. Although significant performance gains are offered by these fine-tuning methods, they are often characterized by high computational costs and a vulnerability to overfitting, particularly in the extreme data-scarce 1-shot regime.

\subsection{Training-Free Caching and Non-parametric Methods}
Training-free adaptation has gained considerable traction due to its ability to perform task-specific refinement without back-propagation. The Tip-Adapter~\cite{zhang2022tip} introduced a non-parametric key-value cache model constructed from few-shot training features, enabling efficient knowledge retrieval at inference time. This baseline was further refined by APE~\cite{zhu2023not}, which introduced adaptive prior refinement to filter discriminative feature channels. More recently, GDA~\cite{wang2024hard} demonstrated that a Gaussian Discriminant Analysis approach, utilizing Mahalanobis distance, provides a strong baseline for training-free adaptation.

\subsection{Kernel Perspectives and Global Regularization}
The theoretical underpinnings of caching methods have recently been scrutinized through a kernel lens. It has been shown in ProKeR~\cite{bendou2025proker} that the adaptation term in Tip-Adapter is mathematically equivalent to a local Nadaraya-Watson (NW) estimator~\cite{nadaraya1964estimating, watson1964smooth}. As local non-parametric methods are inherently biased and lack global task structural information, the ProKeR framework~\cite{bendou2025proker} was proposed to learn a proximal regularizer within a Reproducing Kernel Hilbert Space (RKHS). By formulating the adaptation as a Kernel Ridge Regression (KRR) problem with a global anchor, more robust preservation of prior knowledge is achieved. ReHARK builds upon this global kernel perspective by introducing hybrid semantic-visual priors and multi-scale RBF ensembles, specifically targeting the high-variance challenges of one-shot adaptation.

\begin{figure}[t]
    \centering
    \includegraphics[width=\linewidth]{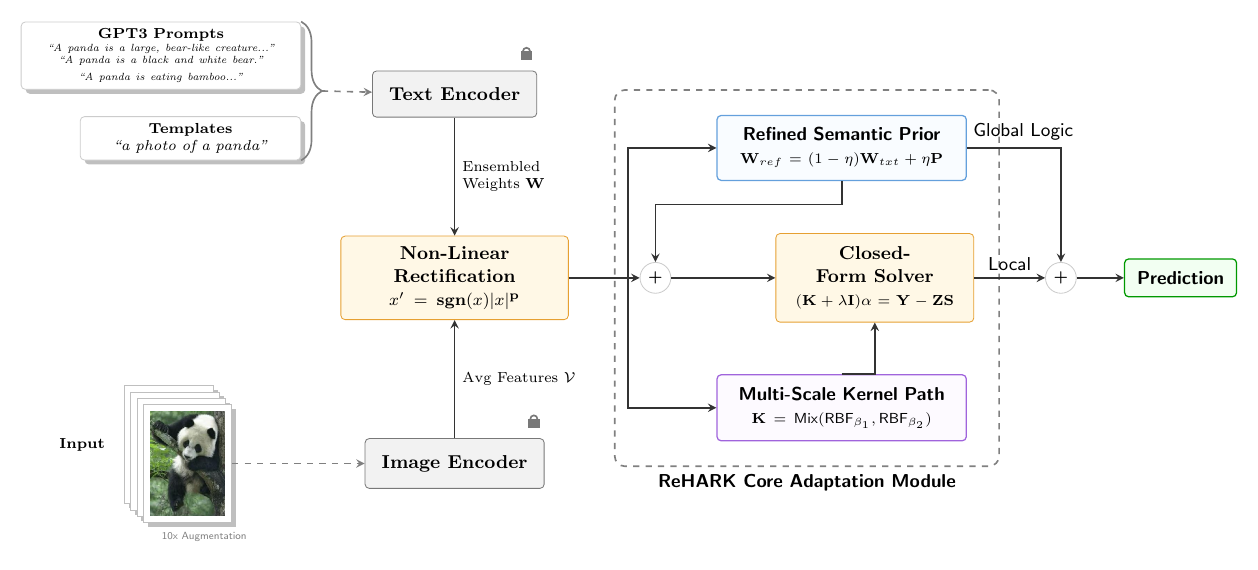}
    \caption{\doublespacing
    The overall architecture of the proposed \textbf{ReHARK} framework. Visual features and ensembled text weights (enriched by GPT3) undergo non-linear rectification before entering the core adaptation module. The system combines a refined semantic prior (global logic) with a multi-scale RBF kernel path (local adaptation) to solve for optimal adaptation coefficients in closed form.}
    \label{fig:architecture}
\end{figure}

\section{Methodology}

In this section, the proposed \textbf{ReHARK} framework is detailed. The architecture is designed to adapt the frozen CLIP backbone to a downstream task using a single visual example per class (1-shot) by utilizing a global kernel regression strategy regularized by hybrid multi-modal priors.

\subsection{Feature Transformation and Rectification}
To mitigate the adverse effects of high-dimensional feature distributions and potential domain shifts, a non-linear power transform is first applied to all visual and textual features~\cite{silva2023closer}. For a given feature vector $\mathbf{x}$, the transformation is defined as:
\begin{equation}
    f(\mathbf{x}, p) = \text{sign}(\mathbf{x}) \cdot |\mathbf{x}|^p
\end{equation}
where $p \in [0.5, 1.0]$ is a learnable scaling factor optimized via a 1000-trial search. This operation is followed by $\ell_2$ normalization to project the features onto a unit hypersphere, ensuring the representations are aligned with the contrastive pre-training objective of the base model.

\subsection{Synergistic Hybrid Prior Construction}
A critical innovation of ReHARK is the construction of a \textbf{Refined Hybrid Prior} that stabilizes the model's global anchor. This is achieved by fusing zero-shot knowledge from CLIP, high-density semantic descriptions from GPT3, and task-specific visual evidence.

First, a \textbf{Base Textual Prior} ($\mathbf{W}_{text}$) is formed by blending CLIP weights ($\mathbf{W}_{clip}$) and GPT3 weights ($\mathbf{W}_{gpt3}$)~\cite{LwEIB-Yang2025}:
\begin{equation}
    \mathbf{W}_{text} = \text{norm}\left( (1 - \gamma) \mathbf{W}_{clip} + \gamma \mathbf{W}_{gpt3} \right)
\end{equation}
where $\gamma$ is a mixing coefficient. Subsequently, this textual prior is refined using visual class prototypes ($\mathbf{P}_{vis}$), which are calculated as the centroids of the available 1-shot visual features:
\begin{equation}
    \mathbf{W}_{prior} = \text{norm}\left( (1 - \omega) \mathbf{W}_{text} + \omega \mathbf{P}_{vis} \right)
\end{equation}
where $\omega$ regulates the balance between pre-trained semantic knowledge and visual evidence.

\subsection{Support Set Augmentation (Bridging)}
To smooth the adaptation manifold in the 1-shot regime, the support set is expanded through a \textbf{Bridge} mechanism. For each visual sample $\mathbf{x}_{vis}$, an intermediate ``bridge'' sample $\mathbf{x}_{bridge}$ is generated by blending the visual feature with its corresponding class-specific refined prior~\cite{bendou2025semobridge}:
\begin{equation}
    \mathbf{x}_{bridge} = \text{norm}\left( \mathbf{x}_{vis} + \eta \mathbf{w}_{label} \right)
\end{equation}
where $\mathbf{w}_{label} \in \mathbf{W}_{prior}$ and $\eta$ is a blending factor. The final augmented support set $\mathbf{S}_{aug}$ is the concatenation of the original visual features and these synthetic bridge samples.

\subsection{Global Proximal Adaptation in RKHS}
The adaptation is formulated as a Kernel Ridge Regression (KRR) problem within an RKHS. Unlike local caching methods such as Tip-Adapter, ReHARK solves for a global weight matrix $\boldsymbol{\alpha}$ that minimizes the following regularized objective~\cite{bendou2025proker}:
\begin{equation}
    \min_{\phi \in \mathcal{H}} \sum_{i=1}^{2NK} || \phi(\mathbf{s}_i) - \mathbf{y}_i ||_2^2 + \lambda || \phi - f_{zs} ||_{\mathcal{H}}^2
\end{equation}
where $f_{zs}$ represents the zero-shot predictor defined by $\mathbf{W}_{prior}$, and $2NK$ is the size of the augmented support set. By the representer theorem, the solution for the adaptation coefficients $\boldsymbol{\alpha}$ is obtained in closed form:
\begin{equation}
    \boldsymbol{\alpha} = (\mathbf{K} + \lambda \mathbf{I})^{-1} (\mathbf{Y} - \hat{\mathbf{Y}}_{zs})
\end{equation}
where $\hat{\mathbf{Y}}_{zs} = \sigma_{zs} (\mathbf{S}_{aug} \mathbf{W}_{prior}^\top)$ represents the zero-shot residuals.

\subsection{Adaptive Multi-Scale RBF Kernels}
To capture feature geometries across diverse scales, an ensemble Multi-Scale RBF kernel is employed. Following the principles of Multiple Kernel Learning (MKL)~\cite{gonen2011multiple}, the kernel $\mathbf{K}(\mathbf{x}, \mathbf{x}')$ is defined as a convex combination of two Gaussian (RBF) kernels~\cite{orr1996introduction} with distinct bandwidths:
\begin{equation}
    \mathbf{K}(\mathbf{x}, \mathbf{x}' ) = \pi \exp\left(-\beta_1 ||\mathbf{x} - \mathbf{x}'||_2^2\right) + (1 - \pi) \exp\left(-\beta_2 ||\mathbf{x} - \mathbf{x}'||_2^2\right)
\end{equation}
where $\beta_1$ and $\beta_2$ capture local and global similarities respectively, and $\pi \in [0, 1]$ is the mixing weight. The final inference for a test query $\mathbf{x}_q$ is computed as the solution to the Proximal Kernel Ridge Regression problem introduced by ProKeR~\cite{bendou2025proker}:
\begin{equation}
    \Phi(\mathbf{x}_q) = \sigma_{zs} (\mathbf{x}_q \mathbf{W}_{prior}^\top) + \mathbf{K}(\mathbf{x}_q, \mathbf{S}_{aug}) \boldsymbol{\alpha}
\end{equation}
where $\sigma_{zs}$ acts as the zero-shot scaling factor and $\boldsymbol{\alpha}$ represents the learned global adaptation coefficients.
\section{Experiments}

\subsection{Datasets and Evaluation Protocol}
The proposed \textbf{ReHARK} framework is evaluated across 11 diverse image classification benchmarks. These datasets encompass a wide variety of domains, including general objects (ImageNet~\cite{deng2009imagenet}, Caltech101~\cite{feifei2004learning}), fine-grained categories (OxfordPets~\cite{parkhi2012cats}, StanfordCars~\cite{krause20133d}, OxfordFlowers~\cite{nilsback2008automated}, Food101~\cite{bossard2014food}, FGVCAircraft~\cite{maji2013fine}), scenes (SUN397~\cite{xiao2010sun}), textures (DTD~\cite{cimpoi2014describing}), satellite imagery (EuroSAT~\cite{helber2019eurosat}), and action recognition (UCF101~\cite{soomro2012ucf101}).

Following the established protocol in the \textbf{CoOp} benchmark~\cite{zhou2022learning}, the evaluation is conducted in the one-shot regime. Hyperparameter selection is performed using validation shots for each specific dataset, after which the optimized configuration is applied to the full test set. This protocol ensures that the reported results accurately reflect the model's ability to adapt to distinct data geometries while utilizing the limited visual evidence available in the few-shot setting.

\subsection{Implementation Details}
The \textbf{ViT-B/16 CLIP} backbone is utilized as the base vision-language model, with the ResNet-50 (RN50) configuration employed for comparative experiments~\cite{radford2021learning}. All computations are performed on a single \textbf{NVIDIA Tesla P100 GPU} (via Kaggle). To ensure computational efficiency during the optimization phase, a batch size of 4096 is utilized for inference.

The hyperparameter search is conducted using the \textbf{Optuna} framework~\cite{akiba2019optuna}, with a total budget of \textbf{1000 trials} allocated per dataset to ensure convergence of the adaptive RBF scales ($\beta_1, \beta_2$), the power transform factor ($p$), and the synergistic prior mixing weights ($\gamma, \omega$). To enrich the semantic representation of the categories, GPT3 based prompts are integrated following the methodology and templates introduced in \textbf{LwEIB}~\cite{LwEIB-Yang2025}. These descriptions are ensembled to form high-density semantic centroids that serve as the foundation for the hybrid prior refinement step.

\subsection{Main Results}
The 1-shot performance of \textbf{ReHARK} is compared against several prominent baselines in Table~\ref{tab:one_shot_results}. ReHARK achieves a new state-of-the-art average accuracy of \textbf{65.83\%}, outperforming Zero-shot CLIP (58.88\%), GDA (62.24\%), Tip-Adapter (62.85\%), and ProKeR (63.77\%). Notably, on the structure-sensitive EuroSAT dataset, ReHARK achieves \textbf{69.19\%}, establishing a substantial lead over the structural-regularized ProKeR (59.75\%).

\begin{table}[htbp]
\centering
\caption{One-shot classification accuracy (\%) comparison across 11 datasets.}
\label{tab:one_shot_results}
\footnotesize 
\setlength{\tabcolsep}{3pt} 
\begin{tabular}{l ccccccccccc c}
\toprule
\multicolumn{1}{c}{Method} & \rotatebox{80}{ImageNet} & \rotatebox{80}{Caltech101} & \rotatebox{80}{DTD} & \rotatebox{80}{EuroSAT} & \rotatebox{80}{FGVCAircraft} & \rotatebox{80}{Food101} & \rotatebox{80}{OxfordFlowers} & \rotatebox{80}{OxfordPets} & \rotatebox{80}{StanfordCars} & \rotatebox{80}{SUN397} & \rotatebox{80}{UCF101} & \rotatebox{80}{\textbf{Average}} \\
\midrule
Zero-Shot CLIP & 60.35 & 85.68 & 42.91 & 36.27 & 17.01 & 77.37 & 66.02 & 85.72 & 55.75 & 58.82 & 61.78 & 58.88 \\
GDA & 60.68 & 87.29 & 46.26 & 58.30 & 17.78 & 77.42 & 72.08 & 85.49 & 56.78 & 59.93 & 62.65 & 62.24 \\
Tip-Adapter & 60.58 & 88.09 & 45.90 & 56.76 & 19.06 & 77.54 & 75.06 & 86.02 & 57.11 & 60.85 & 64.40 & 62.85 \\
ProKeR & 60.60 & 88.17 & 47.99 & 59.75 & 20.65 & 77.40 & 78.85 & 86.44 & 56.79 & 59.66 & 65.13 & 63.77 \\
\midrule
\textbf{ReHARK} & \textbf{61.88} & \textbf{90.13} & \textbf{49.23} & \textbf{69.19} & \textbf{21.45} & \textbf{77.55} & \textbf{80.82} & \textbf{86.34} & \textbf{59.18} & \textbf{63.53} & \textbf{64.83} & \textbf{65.83} \\
\bottomrule
\end{tabular}
\end{table}
\begin{figure}[t]
    \centering
    \includegraphics[width=\linewidth]{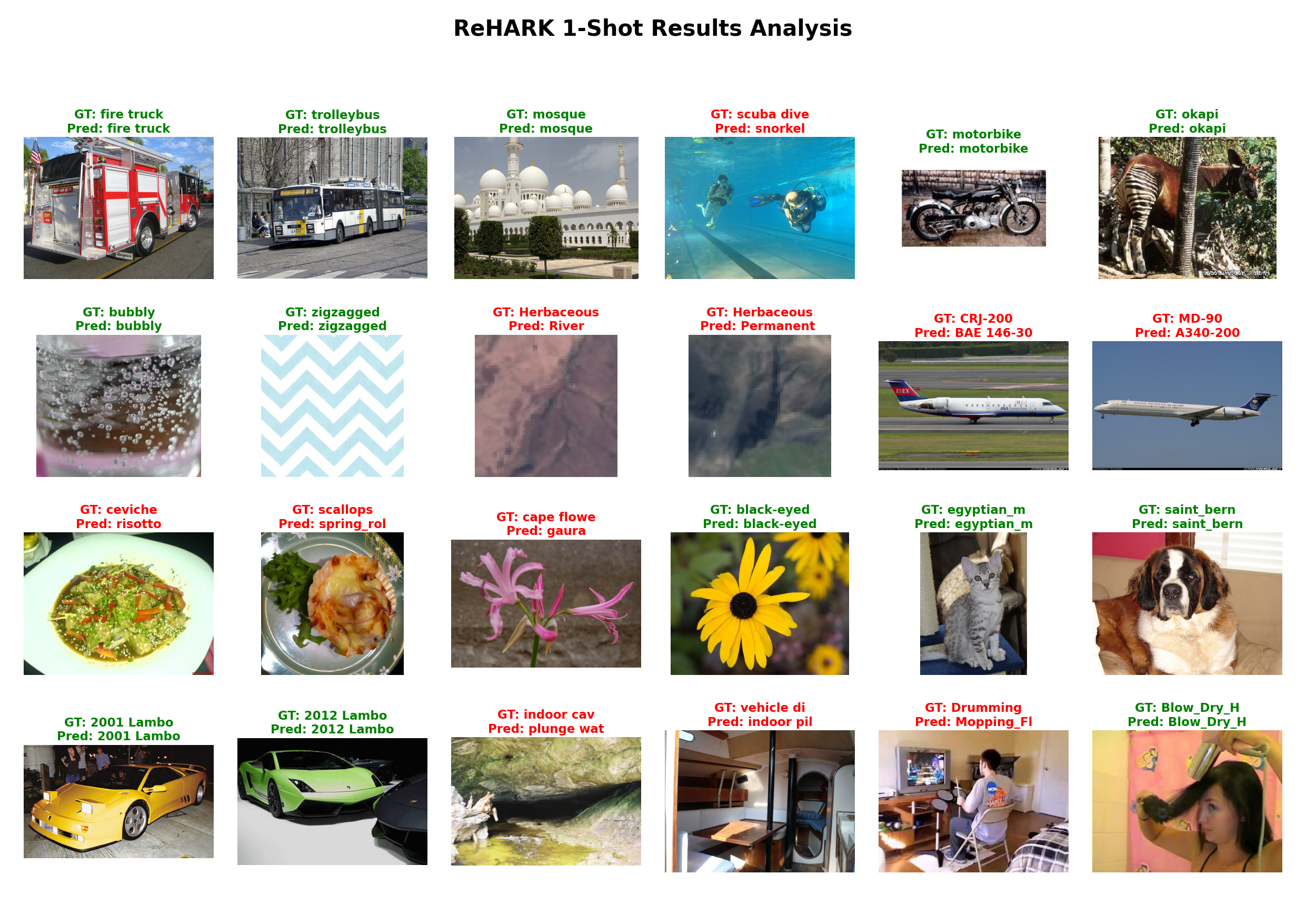}
    \caption{\doublespacing
    Qualitative analysis of ReHARK 1-shot predictions across 11 benchmarks. Green labels indicate correct classifications, while red labels denote misclassifications. The model demonstrates high fidelity in diverse domains, including fine-grained objects and complex scenes.}
    \label{fig:qualitative_results}
\end{figure}

\begin{figure}[t]
    \centering
    \includegraphics[width=\linewidth]{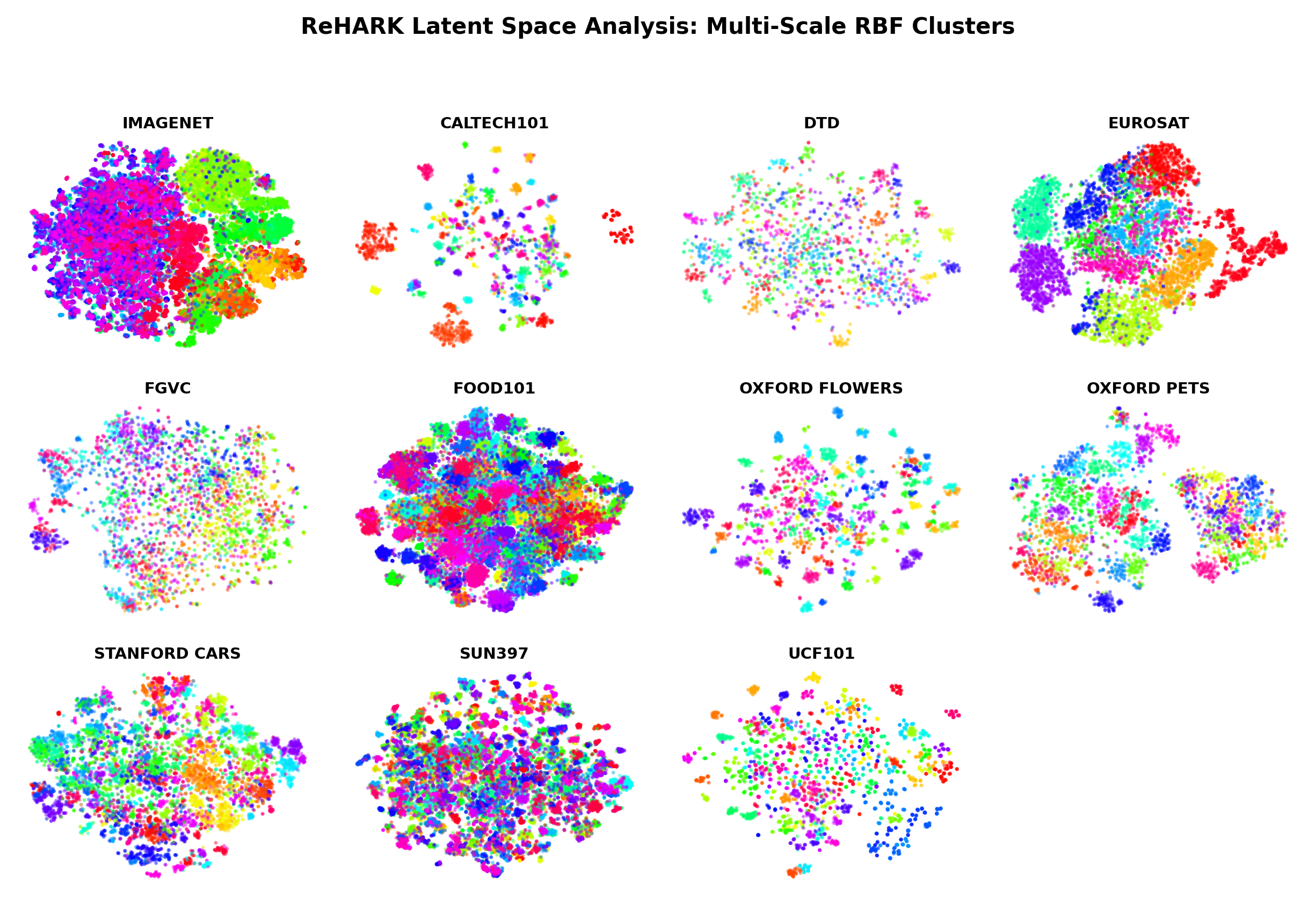}
    \caption{\doublespacing
    t-SNE visualization of the latent space clusters generated by ReHARK. The Multi-Scale RBF kernels effectively capture the local geometry of class distributions across 11 datasets, facilitating distinct separation even with a single support sample per class.}
    \label{fig:tsne_clusters}
\end{figure}

\section{Ablation Study}
In this section, the contribution of each component within the \textbf{ReHARK} framework is systematically evaluated. Unless otherwise specified, experiments are conducted in the 1-shot regime using a ViT-B/16 CLIP backbone.

\subsection{Impact of Architectural Components}
The contribution of each component is evaluated by applying the following mathematical constraints to the optimization objective:

\begin{itemize}
    \item \textbf{NO\_Refine}: $\omega = 0$. The global anchor simplifies to $\mathbf{W}_{\text{prior}} = \mathbf{W}_{\text{text}}$, removing visual prototype influence.
    \item \textbf{NO\_MULTISCALE}: $\pi = 1.0$. The kernel collapses to a single scale: $K(\mathbf{x}, \mathbf{x}') = \exp(-\beta_1 \|\mathbf{x} - \mathbf{x}'\|^2)$.
    \item \textbf{NO\_RECTIFY}: $\eta = 0, \alpha = 0$. Disables moment alignment, resulting in $\hat{\mathbf{x}} = \text{norm}(\mathbf{x})$.
    \item \textbf{NO\_AUGMENT}: $\text{blend\_img} = 0$. Prevents the generation of synthetic bridge samples $\mathbf{x}_{\text{bridge}}$ in the support set.
    \item \textbf{NO\_POWER}: $p = 1.0$. Disables non-linear feature rectification $f(\mathbf{x}, p) = \text{sign}(\mathbf{x}) \cdot |\mathbf{x}|^p$, resulting in a linear pass-through.
\end{itemize}

\begin{table}[htbp]
\centering
\caption{Ablation study of ReHARK components on 1-shot classification average accuracy (\%) (500 trials).}
\label{tab:component_ablation}
\footnotesize 
\setlength{\tabcolsep}{3pt} 
\begin{tabular}{l ccccccccccc c}
\toprule
\multicolumn{1}{c}{Configuration} & \rotatebox{80}{ImageNet} & \rotatebox{80}{Caltech101} & \rotatebox{80}{DTD} & \rotatebox{80}{EuroSAT} & \rotatebox{80}{FGVCAircraft} & \rotatebox{80}{Food101} & \rotatebox{80}{OxfordFlowers} & \rotatebox{80}{OxfordPets} & \rotatebox{80}{StanfordCars} & \rotatebox{80}{SUN397} & \rotatebox{80}{UCF101} & \rotatebox{80}{\textbf{Avg.}} \\
\midrule
NO\_Refine & 61.54 & 89.75 & 49.67 & 68.15 & 21.15 & 77.76 & 80.32 & 85.91 & 58.19 & 63.25 & 64.75 & 65.49 \\
NO\_MULTISCALE & 61.72 & 89.84 & 49.21 & 68.66 & 21.04 & 77.52 & 81.05 & 85.65 & 59.40 & 63.39 & 65.45 & 65.72 \\
NO\_RECTIFY & 61.57 & 89.83 & 48.52 & 66.53 & 21.06 & 77.80 & 81.03 & 86.38 & 58.84 & 63.53 & 64.69 & 65.43 \\
NO\_AUGMENT & 61.60 & 89.93 & 49.27 & 68.75 & 21.25 & 77.76 & 81.25 & 86.07 & 59.06 & 63.55 & 65.12 & 65.78 \\
NO\_POWER & 61.78 & 89.99 & 49.47 & 68.11 & 21.04 & 77.15 & 78.06 & 85.83 & 58.50 & 63.24 & 65.31 & 65.32 \\
\midrule
\textbf{ReHARK} & 62.09 & 90.11 & 49.29 & 68.56 & 21.22 & 77.60 & 80.96 & 86.01 & 58.88 & 63.59 & 64.97 & 65.75 \\
\bottomrule
\end{tabular}
\end{table}
\begin{figure}[t]
    \centering
    \includegraphics[width=0.7\linewidth]{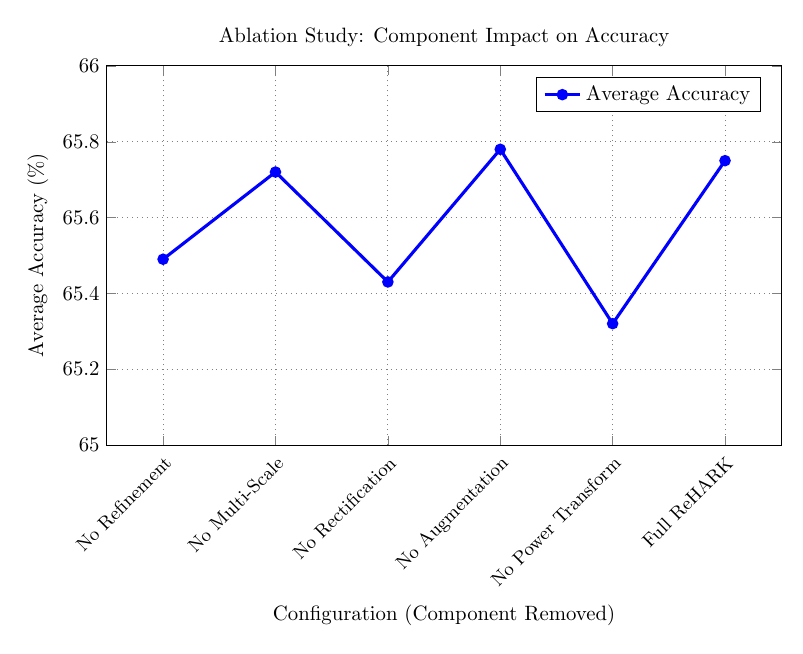}
    \caption{\doublespacing
    Ablation study evaluating the impact of individual architectural components. Removing the \textit{Power Transform} causes the most significant performance degradation ($65.32\%$), while the \textit{Full ReHARK} configuration maintains robust accuracy ($65.75\%$) across all components.}
    \label{fig:architectural_ablation}
\end{figure}
\subsection{Synergistic Prior Modalities}
The synergy between CLIP knowledge, GPT-3 semantic descriptions, and visual evidence is analyzed in Table \ref{tab:modality_ablation}. It is demonstrated that relying solely on visual information (\textit{ONLY\_VISUAL}) yields a drastic performance collapse to \textbf{43.83\%} average accuracy, as a single visual shot provides insufficient coverage of the class distribution. Conversely, the inclusion of GPT-3 descriptors (\textit{ONLY\_TEXT\_GPT}) significantly stabilizes performance (64.32\%), while the full hybrid fusion achieves the highest accuracy of \textbf{65.75\%}.

\begin{table}[htbp]
\centering
\caption{Effect of different modality priors on 1-shot adaptation average accuracy (\%).}
\label{tab:modality_ablation}
\footnotesize 
\setlength{\tabcolsep}{3pt} 
\begin{tabular}{l ccccccccccc c}
\toprule
\multicolumn{1}{c}{Modality} & \rotatebox{80}{ImageNet} & \rotatebox{80}{Caltech101} & \rotatebox{80}{DTD} & \rotatebox{80}{EuroSAT} & \rotatebox{80}{FGVCAircraft} & \rotatebox{80}{Food101} & \rotatebox{80}{OxfordFlowers} & \rotatebox{80}{OxfordPets} & \rotatebox{80}{StanfordCars} & \rotatebox{80}{SUN397} & \rotatebox{80}{UCF101} & \rotatebox{80}{\textbf{Avg.}} \\
\midrule
ONLY\_TEXT\_GPT & 61.06 & 89.70 & 48.72 & 68.63 & 20.25 & 74.79 & 80.16 & 83.44 & 56.98 & 61.55 & 62.27 & 64.32 \\
ONLY\_VISUAL & 25.63 & 76.13 & 35.84 & 63.41 & 16.05 & 38.30 & 67.42 & 42.08 & 30.94 & 38.15 & 48.20 & 43.83 \\
\midrule
\textbf{FULL SYNERGY} & \textbf{62.09} & \textbf{90.11} & \textbf{49.29} & \textbf{68.56} & \textbf{21.22} & \textbf{77.60} & \textbf{80.96} & \textbf{86.01} & \textbf{58.88} & \textbf{63.59} & \textbf{64.97} & \textbf{65.75} \\
\bottomrule
\end{tabular}
\end{table}
\begin{figure}[t]
    \centering
    \includegraphics[width=0.65\linewidth]{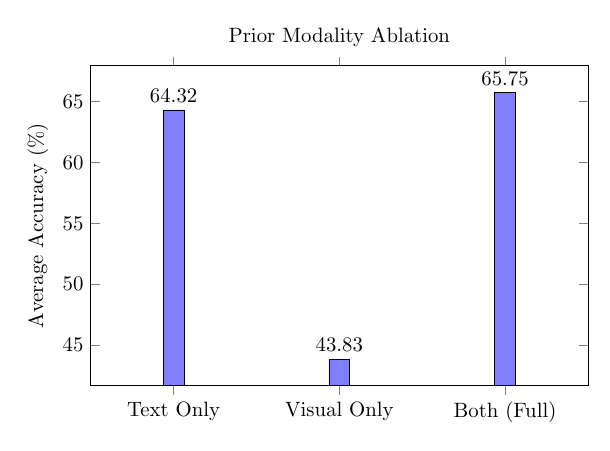}
    \caption{\doublespacing
    Prior modality ablation study. Relying solely on visual priors results in a sharp performance drop to $43.83\%$. The combination of both textual and visual modalities in the full ReHARK framework yields the best average accuracy of $65.75\%$.}
    \label{fig:modality_ablation}
\end{figure}

\subsection{Impact of Search Budget and Kernel Choice}

The robustness of the adaptation process is analyzed along two complementary dimensions: the optimization budget and the kernel function selection.

\paragraph{Optimization Budget}
As the trial budget increases from 50 to 1,000, the average accuracy consistently improves from 64.87\% to a state-of-the-art 65.83\%. This trend, summarized in Table~\ref{tab:trials_ablation}, demonstrates the effectiveness of leveraging the Optuna framework to refine adaptive scales and mixing weights through an expanded search space.
\begin{table}[htbp]
\centering
\caption{Impact of search trials on 1-shot classification accuracy (\%).}
\label{tab:trials_ablation}
\footnotesize 
\setlength{\tabcolsep}{3.5pt} 
\begin{tabular}{l ccccccccccc c}
\toprule
\multicolumn{1}{c}{Trials} & \rotatebox{80}{ImageNet} & \rotatebox{80}{Caltech101} & \rotatebox{80}{DTD} & \rotatebox{80}{EuroSAT} & \rotatebox{80}{FGVCAircraft} & \rotatebox{80}{Food101} & \rotatebox{80}{OxfordFlowers} & \rotatebox{80}{OxfordPets} & \rotatebox{80}{StanfordCars} & \rotatebox{80}{SUN397} & \rotatebox{80}{UCF101} & \rotatebox{80}{\textbf{Avg.}} \\
\midrule
50   & 59.69 & \textbf{90.17} & 49.25 & 67.83 & 20.43 & 76.54 & 79.90 & 85.09 & 57.80 & 62.63 & 64.26 & 64.87 \\
100  & 60.52 & 89.99 & 49.17 & 67.80 & 21.04 & 77.08 & 80.08 & 85.48 & 58.48 & 63.16 & \textbf{65.34} & 65.29 \\
500  & \textbf{62.09} & 90.11 & \textbf{49.29} & 68.56 & 21.22 & \textbf{77.60} & \textbf{80.96} & 86.01 & 58.88 & \textbf{63.59} & 64.97 & 65.75 \\
1000 & 61.88 & 90.13 & 49.23 & \textbf{69.19} & \textbf{21.45} & 77.55 & 80.82 & \textbf{86.34} & \textbf{59.18} & 63.53 & 64.83 & \textbf{65.83} \\
\bottomrule
\end{tabular}
\end{table}
\begin{figure}[t]
    \centering
    \includegraphics[width=0.7\linewidth]{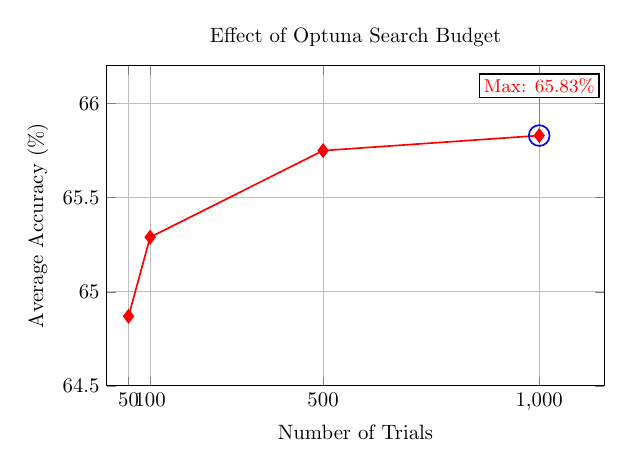}
    \caption{\doublespacing
    Sensitivity analysis of the Optuna search budget. Accuracy increases with the number of trials, reaching a peak of $65.83\%$ at 1,000 trials. Significant gains are observed when increasing the budget from 50 to 500 trials.}
    \label{fig:trail_ablation}
\end{figure}
\paragraph{Kernel Selection}
The adaptation performance is strongly influenced by the choice of the kernel function $K(\mathbf{x}, \mathbf{x}')$ within the global proximal regularization framework. We compare the Radial Basis Function (RBF) kernel against Linear and Laplacian alternatives. The kernel formulations evaluated in Table~\ref{tab:kernel_type_ablation} are defined as follows:

\begin{itemize}
    \item \textbf{Linear Kernel:} Corresponds to the standard dot product in the original feature space:
    \begin{equation}
    K_{\text{Linear}}(\mathbf{x}, \mathbf{x}') = \mathbf{x}^{\top} \mathbf{x}'.
    \end{equation}

    \item \textbf{Laplacian Kernel:} Known for being less smooth than the RBF kernel, it is defined as:
    \begin{equation}
    K_{\text{Laplacian}}(\mathbf{x}, \mathbf{x}') = \exp\left(-\beta \, \lVert \mathbf{x} - \mathbf{x}' \rVert_{1}\right).
    \end{equation}

    \item \textbf{RBF (Gaussian) Kernel:} Captures smooth local similarities:
    \begin{equation}
    K_{\text{RBF}}(\mathbf{x}, \mathbf{x}') = \exp\left(-\beta \, \lVert \mathbf{x} - \mathbf{x}' \rVert_{2}^{2}\right).
    \end{equation}
\end{itemize}

As reported in Table~\ref{tab:kernel_type_ablation}, the RBF kernel achieves the highest average accuracy of 65.83\%, significantly outperforming the Linear (55.45\%) and Laplacian (60.84\%) kernels. This result validates the superior capability of the RBF kernel to project multi-modal features into a high-dimensional space suitable for non-linear adaptation.

\begin{table}[htbp]
\centering
\caption{Impact of different kernel types on 1-shot classification accuracy (\%).}
\label{tab:kernel_type_ablation}
\footnotesize 
\setlength{\tabcolsep}{3pt} 
\begin{tabular}{l ccccccccccc c}
\toprule
\multicolumn{1}{c}{Kernel} & \rotatebox{80}{ImageNet} & \rotatebox{80}{Caltech101} & \rotatebox{80}{DTD} & \rotatebox{80}{EuroSAT} & \rotatebox{80}{FGVCAircraft} & \rotatebox{80}{Food101} & \rotatebox{80}{OxfordFlowers} & \rotatebox{80}{OxfordPets} & \rotatebox{80}{StanfordCars} & \rotatebox{80}{SUN397} & \rotatebox{80}{UCF101} & \rotatebox{80}{\textbf{Avg.}} \\
\midrule
LINEAR    & 26.26 & 88.38 & 47.77 & 63.23 & 18.54 & 45.21 & 77.51 & 80.05 & 44.99 & 55.47 & 62.50 & 55.45 \\
LAPLACIAN & 43.30 & 89.95 & 49.03 & 57.43 & 21.21 & 54.41 & 80.71 & 85.79 & \textbf{59.24} & 63.47 & 64.70 & 60.84 \\
RBF       & \textbf{61.88} & \textbf{90.13} & \textbf{49.23} & \textbf{69.19} & \textbf{21.45} & \textbf{77.55} & \textbf{80.82} & \textbf{86.34} & 59.18 & \textbf{63.53} & \textbf{64.83} & \textbf{65.83} \\
\bottomrule
\end{tabular}
\end{table}
\begin{figure}[t]
    \centering
    \includegraphics[width=0.85\linewidth]{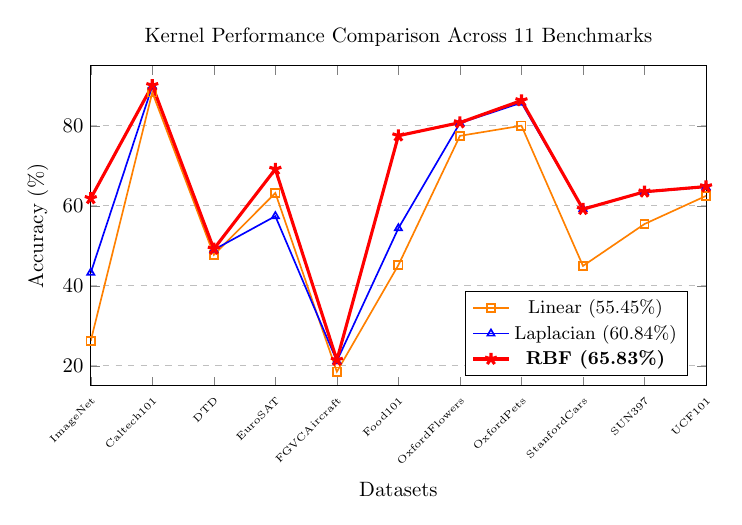}
    \caption{\doublespacing
    Performance comparison of different kernel functions across 11 benchmarks. The RBF kernel (red) consistently achieves the highest accuracy ($65.83\%$), significantly outperforming the Linear ($55.45\%$) and Laplacian ($60.84\%$) baselines, especially on ImageNet and Food101.}
    \label{fig:kernel_ablation}
\end{figure}
\section{Limitations and Future Work}
\label{sec:limitations}

Despite its performance, ReHARK faces several constraints:
\begin{itemize}
    \item \textbf{Search Budget:} The 1,000-trial Optuna search introduces computational overhead during hyperparameter tuning, though inference remains training-free.
    \item \textbf{LLM Dependency:} Generic GPT3 descriptions may lack discriminative power for highly specialized or technical domains.
    \item \textbf{Modality Gap:} High intra-class variance in 1-shot scenarios still presents challenges for visual-textual alignment.
\end{itemize}

\paragraph{Future Work} 
To further advance the capabilities of the proposed framework, several research directions will be pursued. First, \textbf{online hyperparameter prediction} will be explored to eliminate the search phase, thereby streamlining the adaptation process. Additionally, the framework is intended to be \textbf{extended to Large Vision-Language Models (LVLMs)} to leverage their enhanced reasoning and zero-shot capabilities. Finally, \textbf{generative models} will be utilized to create high-fidelity synthetic ``bridge'' samples, aimed at further refining the alignment between textual and visual modalities in few-shot scenarios.

\section{Conclusion}
\label{sec:conclusion}

This paper introduced ReHARK, a training-free framework for one-shot vision-language adaptation. By utilizing global proximal regularization in an RKHS, ReHARK successfully mitigates the boundary biases of local methods. The integration of GPT3 hybrid priors with multi-scale RBF kernels allows for robust feature geometry capture. Experiments across 11 benchmarks demonstrate that Re-HARK establishes a new state-of-the-art with an average accuracy of 65.83\%, particularly excels in structure-sensitive and multi-modal adaptation tasks.

\bibliographystyle{elsarticle-num}
\bibliography{reference}

\end{document}